\pdfoutput=1
\documentclass[11pt]{article}

\usepackage{acl}

\usepackage{times}
\usepackage{latexsym}
\usepackage{amsmath}
\usepackage{colortbl}
\usepackage[T1]{fontenc}
\usepackage[utf8]{inputenc}

\usepackage{microtype}
\usepackage{listings}
\usepackage{color}
\usepackage{colortbl}
\usepackage{multirow}
\usepackage{amssymb}
\usepackage[ruled]{algorithm2e}  
\usepackage{multirow}
\usepackage{booktabs}
\usepackage{adjustbox}
\usepackage{graphicx, subcaption}

\newcommand{\ie}{\textit{i.e.}}
\newcommand{\eg}{\textit{e.g.}}

\usepackage[nameinlink]{cleveref}
\usepackage{enumitem}
\title{Coarse-to-Fine: Hierarchical Multi-task Learning \\ for Natural Language Understanding}

\author{
Zhaoye Fei$^{1,4\dag}$, Yu Tian$^{1\dag}$, Yongkang Wu$^{1\dag}$, Xinyu Zhang$^{1\dag}$, Yutao Zhu$^2$, Zheng Liu$^1$, \\ {\bf Jiawen Wu$^4$, Dejiang Kong$^1$, Ruofei Lai$^1$, Zhao Cao$^{1*}$, Zhicheng Dou$^3$ and Xipeng Qiu$^4$} \\
$^1$Huawei Poisson Lab, China \\
$^2$University of Montreal, Montreal, Quebec, Canada\\
$^3$Gaoling School of Artificial Intelligence, Renmin University of China, Beijing, China \\
$^4$School of Computer Science, Fudan University, Shanghai, China \\
\texttt{\{wuyongkang7,zhangxinyu35,caozhao1\}@huawei.com}
}
        
\begin{document}
\maketitle

\def\thefootnote{\dag}\footnotetext{These authors contributed equally to this work.}
\def\thefootnote{*}\footnotetext{Corresponding author.}
\def\thefootnote{\arabic{footnote}}

\begin{abstract}

Generalized text representations are the foundation of many natural language understanding tasks. To fully utilize the different corpus, it is inevitable that models need to understand the relevance among them. However, many methods ignore the relevance and adopt a single-channel model (a coarse paradigm) directly for all tasks, which lacks enough rationality and interpretation. In addition, some existing works learn downstream tasks by stitches skill block (a fine paradigm), which might cause irrational results due to its redundancy and noise. In this work, we first analyze the task correlation through three different perspectives, \ie, data property, manual design, and model-based relevance, based on which the similar tasks are grouped together. Then, we propose a hierarchical framework with a coarse-to-fine paradigm, with the bottom level shared to all the tasks, the mid-level divided to different groups, and the top-level assigned to each of the tasks. This allows our model to learn basic language properties from all tasks, boost performance on relevant tasks, and reduce the negative impact from irrelevant tasks. Our experiments on 13 benchmark datasets across five natural language understanding tasks demonstrate the superiority of our method.

\end{abstract}

\section{Introduction}
% \textit{``It has been said that something as small as the flutter of a butterfly's wing can ultimately cause a typhoon halfway around the world.''}
% {\rightline{--- Chaos Theory}}
Pre-trained language models have achieved great success on various natural language processing (NLP) tasks. Meanwhile, \textit{pre-train-then-fine-tuning} has gradually become the mainstream paradigm~\cite{bert,liu2019roberta,yang2019xlnet}. The pre-training process aims to learn a general language representation from the large-scale corpus. Such representations can be further fine-tuned on downstream datasets and perform specific tasks. Though great performance has been achieved, it is costly to fine-tune and save independent representation for each task. Therefore, researchers propose several multi-task learning methods~\cite{phang2018sentence,liu2019multi,clark2019bam} using a single-channel model (a coarse paradigm) to solve multiple tasks.

% \begin{figure*}[t]
% 	\begin{center}
% 		\includegraphics[width=1\linewidth]{fig/structure.pdf}
% 	\end{center}
% \caption{the architecture of single-channel framework, coarse-to-fine framework (Our model) and dense multi-channel framework. Compared to other methods, our approach takes reasonable account of task relevance and the trade-off between performance and parameters.}
% \label{fig:compare}
% \end{figure*}
\begin{figure*}
    \centering
    \includegraphics[width=0.9\linewidth]{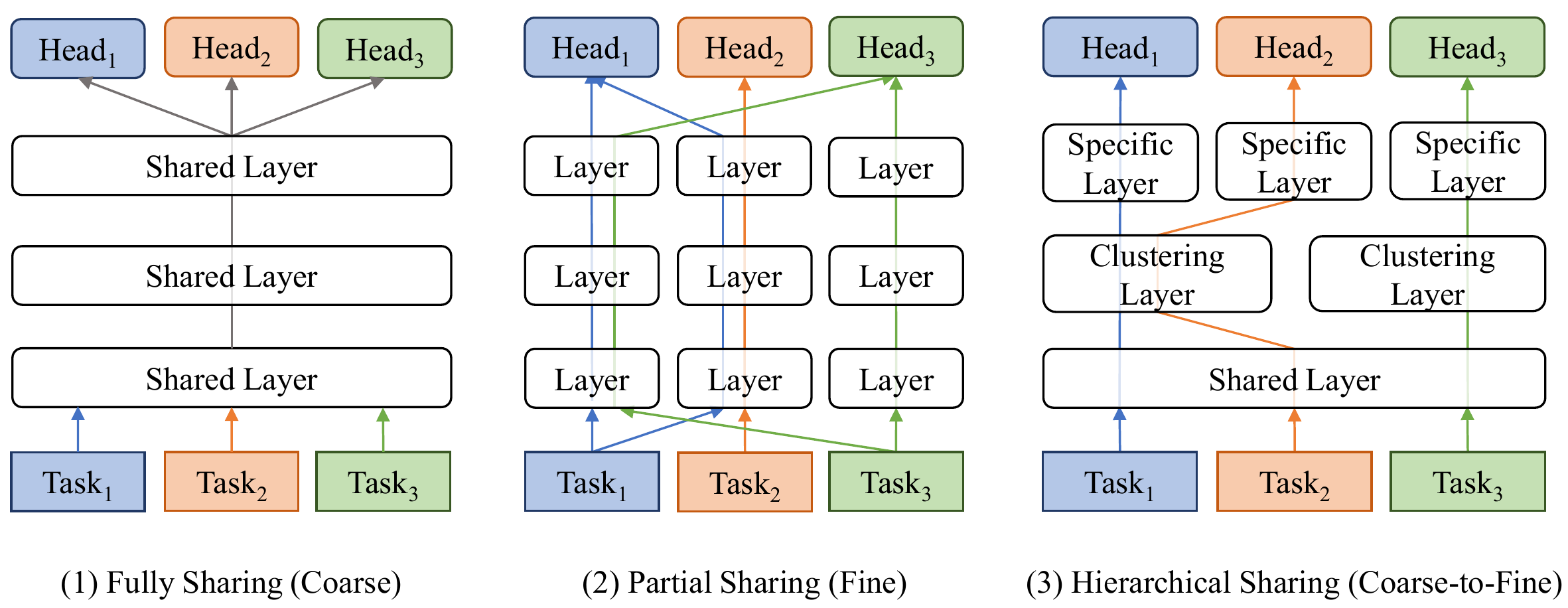}
    \caption{The illustration of hard sharing structure, partial sharing structure, and our hierarchical sharing structure. The three tasks and the corresponding activated paths are in different colors. In our approach, the first and second are two relevant tasks, so they share the same task-clustering layers. Finally, they are fed into different task-specific layers.}
    \label{fig:structure}
    \vspace{-10px}
\end{figure*}

Inspired by human learning, multi-task learning believes that tasks can interact and boost each other. Therefore, to obtain a more robust representation that can handle a variety of tasks, a straightforward idea is to fine-tune a pre-trained model on many tasks simultaneously. Numerous previous studies have been conducted along this path. For example, MT-DNN~\cite{liu2019multi} uses a transformer-based model as a shared encoder and trains it on multiple downstream tasks (as shown in the left side of Figure~\ref{fig:structure}). By this coarse paradigm, the representation model can be more generalized and robust. Although it has been observed that some combinations of tasks yield improvements, this is not always the case. Researchers~\cite{aribandi2021ext5} have also found that the impact between tasks is a double-edged sword, meaning that some tasks may also hurt others. Indeed, modeling heterogeneous tasks often require distinct representation spaces. For example, naively training natural language inference (NLI) tasks with different hypothesis types together can lead to performance declines.

Some recent work proposes sparsely activating multiple modules for different tasks~\cite{zhang2022skillnetnlu} to mitigate the negative effects across tasks. These fine paradigms activate distinct modules according to predefined skills for learning downstream tasks (as shown in the middle part of Figure~\ref{fig:structure}). However, the number of parameters is also multiplied when multiple modules are activated for a task. This problem becomes even more severe when large pre-trained language models are applied.
%To mitigate the negative effects across tasks, some recent work proposes sparsely activating multiple modules for different tasks~\cite{zhang2022skillnetnlu}. These fine paradigms activate distinct modules according to predefined skills for learning downstream tasks (as shown in the middle part of Figure~\ref{fig:structure}). However, the number of parameters is also multiplied when multiple modules are activated for a task. This problem becomes even more severe when large pre-trained language models are applied.

% \begin{figure}[t]
% 	\begin{center}
% 		\includegraphics[width=0.8\linewidth]{fig/overview.eps}
% 	\end{center}
% \caption{the architecture of single-channel framework, coarse-to-fine framework (Our model) and dense multi-channel framework. Compared to other methods, our approach takes reasonable account of task relevance and the trade-off between performance and parameters.
% }
% \label{fig:architecture}
% \end{figure}

To address the aforementioned problems, we first conduct a correlation analysis on four natural language understanding (NLU) categories (a total of 13 datasets) and find that some tasks can complement one another, while others cannot. 
Based on these observations, we propose three measures (\ie, data property, manual design, and model-based relevance) to categorize the tasks into different groups.

Then, inspired by recent studies~\cite{kovaleva-etal-2019-revealing,rogers-etal-2020-primer} that different layers learn information in different levels, we design a novel hierarchical sharing framework, dubbed HMNet, a coarse-to-fine multi-task learning paradigm for natural language understanding (as shown in the right side of Figure~\ref{fig:structure}). Our method is a coarse-to-fine paradigm, where the layers are divided from the bottom up into shared, task-clustering, and task-specific levels. We design the bottom layers as shared, which are optimized by all tasks. Thereafter, we employ distinct layers for different groups obtained in the previous step. In this way, the relevant tasks grouped in the same cluster can boost each other, while the negative impact from tasks in other groups can be avoided. The top layers are totally separated for distinct tasks, so that some task-specific information can be well-captured. It is worth noting that each task only activates a single module in each layer, so our model does not require additional parameters during inference.
Extensive experiments on 13 datasets show that our method can achieve better performance on most NLU tasks, demonstrating its superiority over existing hard sharing or partial sharing structures. 

Our main contributions are three-fold: 

(1) We perform a series of correlation analyses (\ie, data property, manual design, and model-based relevance) on 13 NLU tasks, which sheds light on the positive and negative effects among the different tasks. 

(2) We design a novel hierarchical sharing framework, dubbed HMNet, a coarse-to-fine multi-task learning paradigm for natural language understanding. It can better leverage the positive interactions among different tasks while reducing the negative influence.
% noise.

(3) We conduct experiments on 13 commonly used NLU datasets and validate the effectiveness of our proposed method. The results demonstrate that our framework achieves highly competitive performance while saving over $34\%$ parameters than the partial sharing structure.
% demonstrate that our approach significantly outperforms on most NLU tasks. 

\section{Related Work}
In this section, we briefly introduce some Transformer-based pre-trained models and recent work on multi-task learning.

\subsection{Transformer-based Pre-trained Models}
Transformer is a neural structure consisting of multiple stacked self-attention modules~\cite{bert,liu2019roberta,yang2019xlnet}. With the bidirectional attention mechanism, the model can capture contextual information from both sides effectively. BERT~\cite{bert} proposes a masked language modeling objective to pre-train a Transformer encoder, and achieves dramatic performance on several natural language understanding tasks. Since then, the \textit{pre-train-then-fine-tuning} paradigm has gradually become mainstream. Researchers propose various new pre-training strategies to facilitate the model in several aspects~\cite{yang2019xlnet,bart,t5}. For example, in order to obtain a more robust model, RoBERTa~\cite{liu2019roberta} adapts several tailored training strategies and employs more data. There are also many methods extending the pre-training on Transformer decoders. GPT-2~\cite{gpt2} is a decoder-only structure that is pre-trained in accordance with the language generation objectives. It also performs exceptionally well on many text generation tasks. Though pre-trained models can be fine-tuned for distinct tasks, it is costly to maintain a separate model for each task, especially when the model is huge. 

\subsection{Multi-task Learning} 
Multi-task learning is an integrated learning method in which multiple tasks share the same structure for training simultaneously. It can enhance the generalization and performance of each task. Consequently, multi-task learning can also be applied to pre-trained language models. A typical practice is to fine-tune the pre-trained language models by conducting multiple tasks concurrently~\cite{liu2019multi}. It is reported, however, that not all tasks can boost each other, and that noise may also be introduced~\cite{aribandi2021ext5}. Therefore, it is essential to analyze the relationship between tasks before training them jointly. Some recent work proposes addressing this issue by sparsely activating distinct modules for different tasks~\cite{zhang2022skillnetnlu}. This allows the modules to be trained by relevant tasks. Unfortunately, these approaches usually rely on predefined activation paths, and they have to activate multiple modules in order to achieve high performance during the inference stage. This undoubtedly increases the delay and cost of model application.

The main differences between our method and others are: (1) We analyze the task correlation, and use it to group the tasks. Relevant tasks within the same group will be used to fine-tune the same modules, while irrelevant tasks will not interfere with one another. (2) Our model only activates a single module in each layer for each task, therefore it has the same number of parameters as a single model.

\section{Proposed Method}\label{sec:method}
\subsection{Overview}
The overview of our method is illustrated in the right side of Figure~\ref{fig:structure}. In general, there are three different kinds of layers in our structure: shared layers, task-clustering layers, and task-specific layers. The shared layers are in the bottom and optimized by all tasks. The middle part is the task-clustering layers. Tasks that are classified as relevant (introduced in the next section) will optimize the same group of layers. The top layers are task-specific, meaning that each task has its own module.

Our method adopts a coarse-to-fine paradigm. In this manner, the model can obtain generalized text representations in shared layers (a coarse-grained manner) and effectively interact and learn according to task correlation in the task-clustering layer (a fined-grained manner), then send it to the task-specific layer for specific task training.

\subsection{Task Relevance Analysis}\label{sec:task_relevance}

\begin{figure*}[t]
    \centering
    \begin{subfigure}[b]{0.45\textwidth}
    \includegraphics[width=\textwidth]{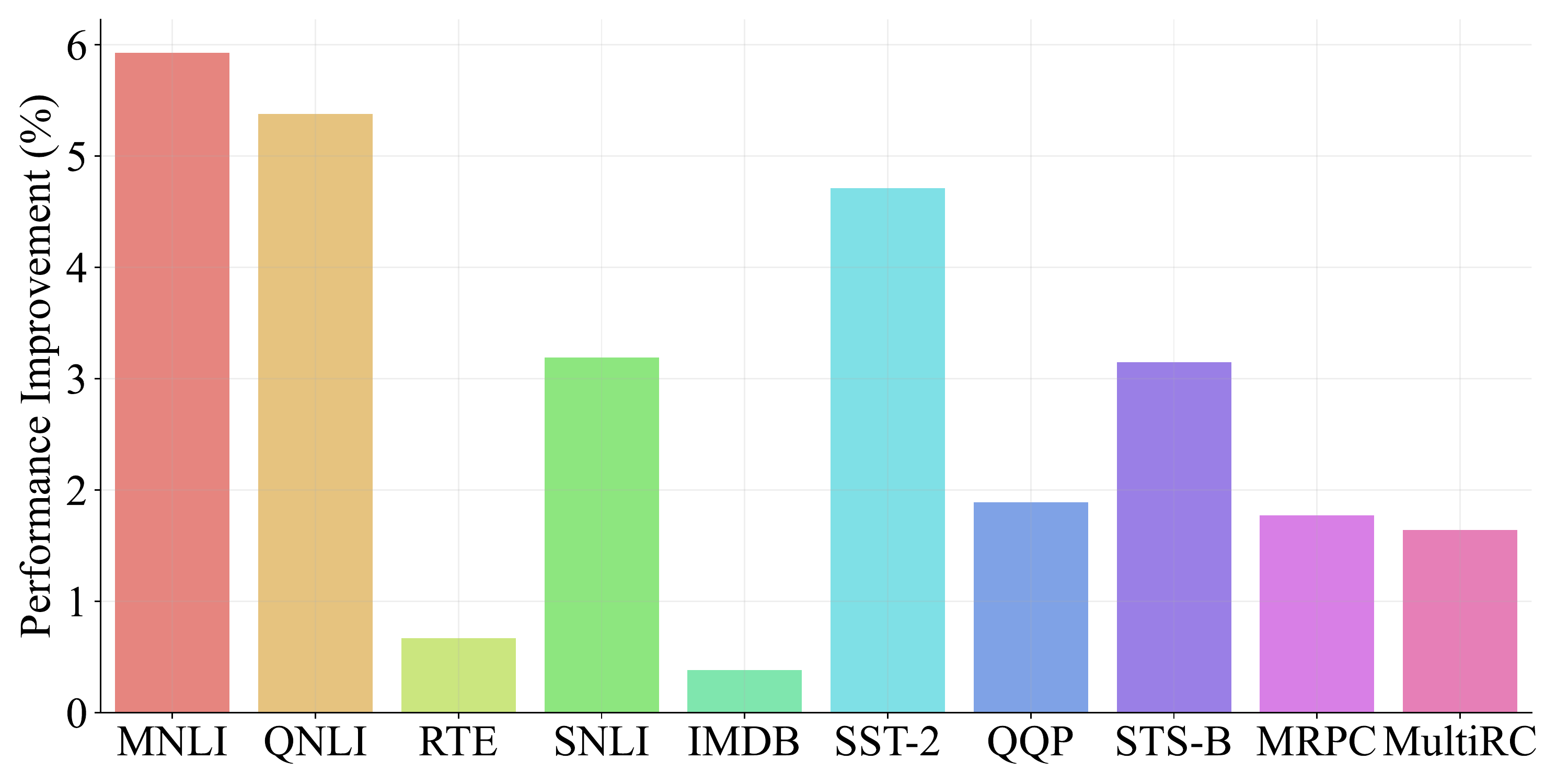}
    % \caption{}
    \end{subfigure}
    \begin{subfigure}[b]{0.45\textwidth}
    \includegraphics[width=\textwidth]{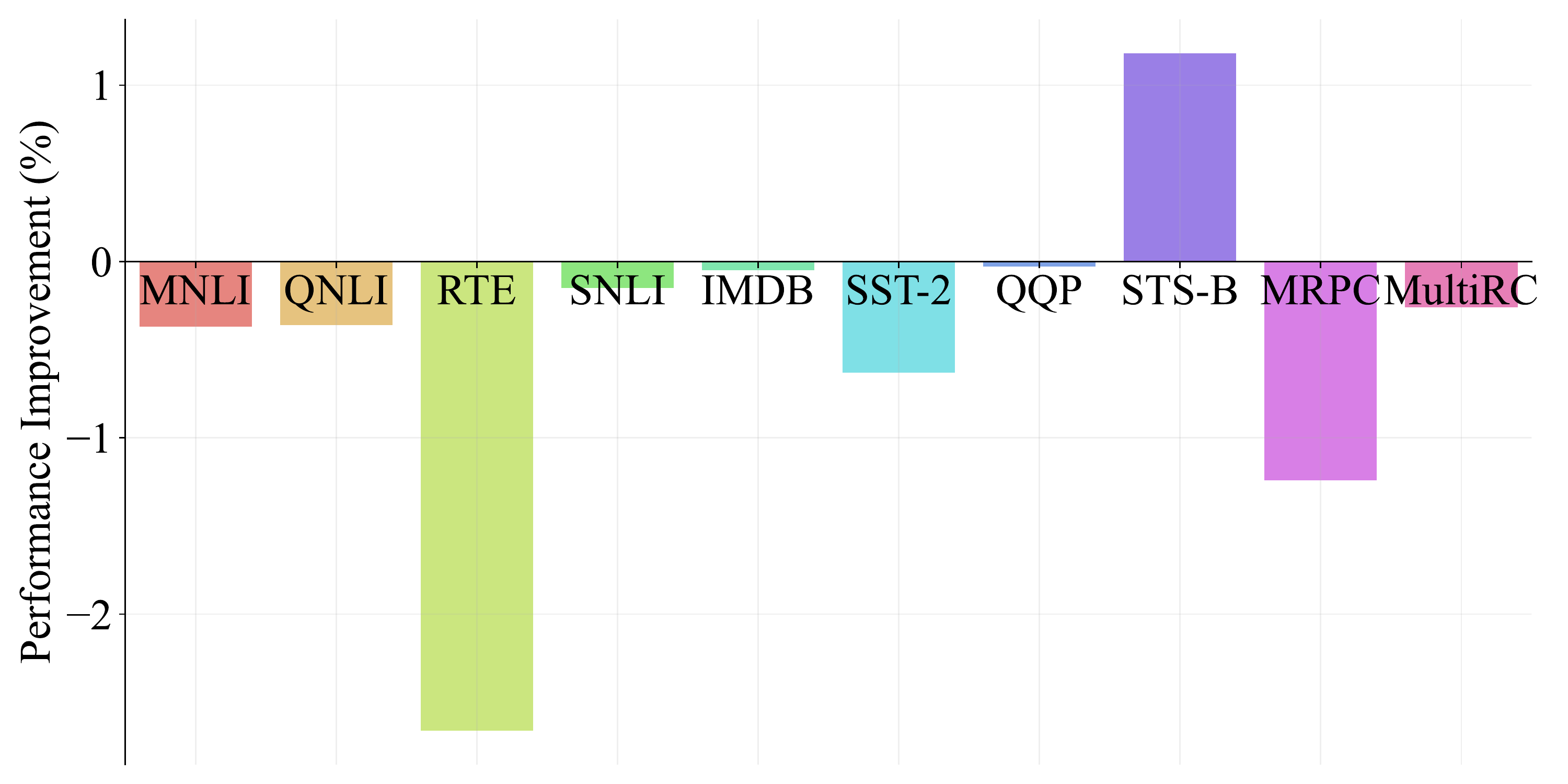}
    % \caption{}
    \end{subfigure}
    \caption{Influence of other tasks on the BoolQ dataset (left) and influence of the BoolQ dataset on others (right).}
    \label{fig:boolq_mcor}
    \vspace{-10px}
\end{figure*}

Existing multi-task learning framework assumes that all tasks can facilitate each other. However, researchers have also reported the negative effect caused by irrelevant tasks~\cite{aribandi2021ext5}. To investigate the influence of each task, we devise three methods from different perspectives to measure the task relevance.

\paragraph{Data Property}
Since all tasks we study are based on natural language texts, we first analyze the relevance between tasks through the property of their data. Specifically, we consider the syntactic information and employ vocabulary co-occurrence to measure task relevance. Formally, given datasets of two tasks denoted as $\mathcal{D}_\text{s}$ and $\mathcal{D}_\text{t}$, where $s$ and $t$ stand for \textit{source} and \textit{target}, we compute the vocabulary co-occurrence as follows:
\begin{equation}
% \begin{split}
    r_{\text{cs}} = |{V}_{\text{s}} \cap {V}_{\text{t}}| / |{V}_{\text{s}}|, \quad
    r_{\text{ct}} = |{V}_{\text{s}} \cap {V}_{\text{t}}| / |{V}_{\text{t}}|,
    \label{eq:text_corr}
% \end{split}
\end{equation}
where $V_\text{s}$ and $V_\text{t}$ are the vocabulary of the source and target datasets, respectively, $r_{\text{cs}}$ measures the ratio of words in the source dataset that are shared with the target datasets, and $r_{\text{ct}}$ represents the opposite. Intuitively, a higher value of $r_{\text{cs}}$ indicates a higher impact of the target task on the source task. Note that this metric also considers the data size, as a larger dataset usually contains more words. For example, MNLI~\cite{williams2018broad}, as the largest dataset, has a large vocabulary co-occurrence with others. It can provide some basic language knowledge for other tasks. Conversely, WNLI~\cite{wang-etal-2018-glue} is a small dataset with only hundreds of samples, and it may have limited impacts on other tasks.
Due to the space limitation, we show the entire results in Appendix~\ref{app:task_relevance}. 
% Table~\ref{tab:text_corr} shows part of the results of the relevance between tasks. 

% \paragraph{Human Observation}
% Inspired by a recent work that designs various paths for different tasks~\cite{zhang2022skillnetnlu}, we also examine the tasks artificially. We discover that whereas these tasks are collected from diverse domains and designed for unique purposes, their input formats have some similarities. Based on this observation, we divide them into three groups, \ie, single sentence, paired sentence, and question answering. We assume tasks having similar input format require similar capabilities of models. For instance, the task with the input of paired sentence generally relies on models' judgment on the relationship between the pairs, while question answering tasks ask the model to inference the answer from the paragraph according to the question. As a result, we have three groups: \{\}, \{\}, and \{\}. 

\paragraph{Manual Design} 
Inspired by a recent work that designs various paths for different tasks~\cite{zhang2022skillnetnlu}, we also examine the tasks artificially. Based on the manually designed task purpose, we divide them into four groups. \ie, Natural language inference, sentiment classification, similarity and paragraphing, and question answering. We assume tasks that have the same purpose supply the same systematic information for models. For instance, the NLI tasks generally rely on the models' judgment according to the deep semantics information between premise and hypothesis, while the question answering tasks ask the model to infer the answer from the passage according to the question. These tasks and their category will be reported in Section~\ref{sec:data_metric}.

\paragraph{Model-based Relevance}
In addition, to manually analyze the task or data, we further propose measuring task relevance using neural networks. Previous studies have demonstrated the advantages of multi-task learning~\cite{liu2019multi}, notably that relevant tasks can enhance each other's performance. Consequently, task relevance can be inferred by comparing the performance difference between training two tasks independently and training them jointly. Concretely, for a source and a target task, we fine-tune two models (we use BERT in our experiments) respectively for them. Their average performance is denoted as $f_\text{s}$ and $f_\text{t}$. Afterward, we fine-tune another model (BERT) for the two tasks through standard multi-task learning. The corresponding performance is denoted as $f_{\text{js}}$ and $f_{\text{jt}}$. Finally, we can compute the task relevance as the performance improvement:
\begin{equation}
% \begin{split}
    r_{\text{ms}} = (f_{\text{js}} - f_\text{s}) / f_\text{s}, \quad r_{\text{mt}} = (f_{\text{jt}} - f_\text{t}) / f_\text{t},
    \label{eq:task_corr}
% \end{split}
\end{equation}
% where $r_{ms}$ reflects the influence of the target task on the source task, while $r_{mt}$ represents the opposite. We calculate the relevance for all tasks, and the results about BoolQ as an example in Figure~\ref{fig:boolq_mcor} to highlight the asymmetrical relevance between tasks. The complete form reported in Appendix~\ref{app:task_relevance}.
% % As an example, Figure~\ref{fig:boolq_mcor} shows the task relevance computed by our proposed model-based relevance about BoolQ. 
% In this figure, we find that most tasks have a positive impact on BoolQ, even though the data property and the purpose of these tasks are very different. BoolQ, on the other hand, harms other tasks. 
where $r_{\text{ms}}$ reflects the influence of the target task on the source task, while $r_{\text{mt}}$ represents the opposite. We calculate the relevance for all tasks, and the results about BoolQ as an example in Figure~\ref{fig:boolq_mcor} to highlight the asymmetrical relevance between tasks. The complete form is shown in Appendix~\ref{app:task_relevance}. In Figure~\ref{fig:boolq_mcor}, we find that most tasks have a positive impact on BoolQ, even though the data property and the purpose of these tasks are very different. BoolQ, on the other hand, harms other tasks. One possible reason is learning for QA needs more complicated information and learning on other tasks can supply this information to improve the performance of BoolQ. But as for other tasks, the information obtained by learning in BoolQ like noise damages the learning of themselves. 

Since the asymmetrical relevance between task pairs, we consider the fully shared model is harmful to some tasks in multi-task learning. Inspired by~\citeauthor{rogers-etal-2020-primer}, we consider that the improvement of word-level information training in the low layer occurs a positive impact mostly, whereas the more negative impact is sourced from the damage of the deep semantic learning in the middle and top layers. As a result, we propose a hierarchical sharing method for multi-task learning. 
% The bottom layer of the model captures the general information contained in all tasks, the mid-layer learns more semantic information from the same task category grouped by task relevance, and the top layer learns the task-specific information from each task. 
It is based on a hierarchical sharing of the task with different relevance to leverage more positive interaction and reduce the negative impact of multi-task learning. More detail about the structure is described in the next section.

% Based on these findings, we determine fully shared

% we have a simple hypothesis that the improvement of word-level semantic information training in the low layer from other tasks occurs a positive impact, whereas the damage of deep semantic learning in the mid and top layers is expressed as a negative impact. As a result, we created a hierarchical multi-task architecture to leverage positive interactions and reducing the negative interactions.

% Furthermore, we can determine that the impact between task pairs is asymmetric. So, it's difficult for humans to cluster these tasks to get the best results and these influence cannot reflected by naive data properties clearly, indicating the potential superiority of model-based relevance.  

% The complete form reported in Appendix~\ref{app:task_relevance}. 

% This is also an asymmetrical Figure~\ref{} shows the task relevance computed by our proposed model-based similarity. 

% Interestingly, we find that SST-2 has negative impact on IMDB. They are both for sentiment classification task, from similar domains, and have identical input format. After carefully checking the experimental results, we determine that ... This difference is difficult for humans to detect and cannot be reflected by naive data properties, indicating the potential superiority of model-based similarity. 

\subsection{HMNet}
We propose a hierarchical multi-task learning framework (HMNet) based on task relevance. It is built on Transformer~\cite{vaswani2017attention}, which has been widely applied in NLP tasks.\footnote{Other structures, such as convolutional or recurrent networks, may also be compatible with our method. This is left for future work.} We omit the details of Transformer and refer readers to the original paper. As shown in the right side of Figure~\ref{fig:structure}, HMNet has three different kinds of Transformer layers from the bottom up. All tasks will first pass the shared layers, then, each task will go through task-clustering layers according to their task group (introduced later). Finally, each task has its own task-specific layer, and the associated head is used to accomplish the task.

In the following paragraphs, we first introduce the details of the layers, followed by the task clustering process.

\noindent\textbf{Shared layers:} The shared layers are stacked in the bottom of HMNet. We design bottom layers as shared since it has been demonstrated that they capture low-level semantic or structural information in existing pre-trained models~\cite{jawahar2019does}. We believe that such information is universally contained in all texts, so all tasks collaborate to optimize the shared layers. 

\noindent\textbf{Task-clustering layers:}
Based on our task relevance analysis, we cluster the tasks into various groups. For tasks within the same cluster, the same (set of) clustering layers will be optimized. As illustrated on the right side of Figure~\ref{fig:structure}, the first two tasks are grouped into one cluster, so they optimize the first set of clustering layers. On the contrary, only the third task tunes the second set of clustering layers. With the task-clustering layers, relevant tasks can optimize the same set of parameters, enabling the sharing of their knowledge. In the meantime, the irrelevant tasks can be isolated, thereby eliminating the noise.

\noindent\textbf{Task-specific layers:} 
Task-specific layers are in the top of our HMNet. According to recent studies~\cite{jawahar2019does}, the top layers of pre-trained language models typically learn task-specific knowledge. Therefore, we separate all tasks and let them have their own Transformer layers. There is also a head associated with each task-specific layer to accomplish the task. For example, a linear classifier with an activation function is often applied for classification tasks. 

\paragraph{Clustering Process}
In our task relevance analysis, we propose three methods to measure the task relevance. Among them, the manual design can directly group the tasks into different clusters. For the other two methods, we employ an unsupervised clustering method, \ie, $k$-means. Particularly, given a source task $S$, and $n$ target tasks $\{T_1,\cdots,T_n\}$, we can compute the relevance scores $\{r_{\text{ST}_1},\cdots,r_{\text{ST}_n}\}$ by Equation~(\ref{eq:text_corr}) or~(\ref{eq:task_corr}). Then, we treat the $n$ scores as features and apply $k$-means clustering algorithm. Finally, the tasks can be grouped into $k$ clusters. The grouping results 
are given in the Appendix~\ref{app:task_relevance}.

\paragraph{Remark}
Different from traditional multi-task learning that shares all layers among all tasks, our HMNet only shares layers at the bottom (a coarse-grained manner). The higher layers are shared by only relevant tasks or used alone(a fined-grained manner). Since the layers close to specific tasks are separated, our method can alleviate the contradiction between tasks. On the other hand, each task only activates one channel at each layer, so the total parameters are comparable with a single-channel model. This is much more efficient than sparsely activated structures. 

\subsection{Optimization}
HMNet is based on a multi-layer Transformer, which is similar to existing pre-trained language models, such as BERT~\cite{bert}. Therefore, we use BERT as the backbone model to initialize the parameters in each layer. Then, each task has its own path (as described earlier), and we can use it to fine-tune the corresponding layers. 
The training process our method is summarized in Algorithm~\ref{algor1}. In each epoch, a mini-batch $b_{t}$ is packed, and the HMNet is updated by the path for the dataset $\mathcal{D}_i$.

\begin{algorithm}[t]
    \small
    \caption{Training Process}
    Initialize model parameters from pre-trained BERT; \\
    Set the max number of training epoch $E_m$; \\
    % \KwIn{Datasets $\mathcal{D}_1, \mathcal{D}_2, \cdots, \mathcal{D}_N$ for $N$ tasks}
    \textbf{{Prepare data}} \\
    \For{$i$ in $1,2,\cdots,N$} 
    { 
        Pack the dataset $\mathcal{D}_i$ into minibatch $B_i$;
    } 
    \textbf{Multi-task learning} \\
    \For{{\rm epoch} $t$ {\rm in} $1,2,\cdots,E_m$} 
    {
        Merge all the datasets: $B=B_1\cup B_2 \cdots \cup B_N $; \\
        Shuffle $B$; \\
    \For{$b_{t}$ in $B$}
    {
        // \textit{$b_t$ is a mini-batch of task $t$}; \\
        Feed $b_{t}$ into shared layers $\rightarrow$ $\mathbf{b}_{t}^{s}$;\\
        Feed $\mathbf{b}_{t}^{s}$ into task-clustering layers $\rightarrow$ $\mathbf{b}_{t}^{c}$; \\
        Feed $\mathbf{b}_{t}^{c}$ into task-specific layers $\rightarrow$ $\mathbf{b}_{t}^{t}$; \\
        Feed $\mathbf{b}_{t}^{t}$ into task-specific head; \\
        Compute loss and gradient; \\
        Update model;
    }
    }
    
\label{algor1}
\end{algorithm}

\begin{table*}[t!]
    \centering
    \small
    \begin{tabular}{lccccccc}
    \toprule
         & Metric & Single-task & Multi-task & SkillNet & HMNet$_{\rm d}$ & HMNet$_{\rm md}$ & HMNet$_{\rm m}$ \\
    \midrule
        \multicolumn{8}{l}{\textsc{\textbf{Natural Language Inference}}} \\
        MNLI (m/mm) & Acc. & \textbf{85.0}/84.6 & 84.5/84.8 & 84.5/84.6 & 84.8/84.6 & 84.7/\textbf{85.2} & 84.8/84.9 \\
        QNLI & Acc. & \textbf{91.6} & 90.9 & 90.7 & 90.8 & 91.0 & 91.1 \\
        RTE & Acc. & 66.1 & 79.7 & 77.6 & 73.3 & 79.4 & \textbf{81.2} \\
        WNLI & Acc. & 56.3 & 56.3 & 56.3 & 56.3 & 56.3 & 56.3 \\
        CB & Acc. & 67.8 & 80.3 & 85.7 & \textbf{87.5} & 82.1 & 82.1 \\
        SNLI & Acc. & 91.0 & 91.3 & 91.1 & 91.0 & \textbf{91.4} & 91.1 \\
        %\textit{NLI Average} & Acc. & 77.5 & 81.1 & 81.5 & 81.2 & 81.5 & \textbf{81.6 }\\
        \midrule
        \multicolumn{8}{l}{\textbf{\textsc{Sentiment Classification}}} \\
        IMDB & Acc. & 93.9 & 93.9 & \textbf{94.1} & 93.8 & 93.9 & 93.9 \\
        SST-2 & Acc. & 92.3 & 93.1 & 92.5 & \textbf{93.2} & 92.3 & 93.0 \\
        \midrule
        \multicolumn{8}{l}{\textbf{\textsc{Similarity and Paraphrase}}} \\
        QQP & Acc. & 90.9 & 90.7 & \textbf{91.0} & 90.9 & \textbf{91.0} & 90.9 \\
        STS-B & Corr.  & 85.8 & 85.5 & 86.0 & 86.9 & \textbf{87.5} & 87.4 \\
        % \midrule
        % \multicolumn{7}{l}{\textsc{\textbf{Paragraphing}}} \\
        MRPC & Acc. & 83.3 & 81.4 & 85.7 & 88.2 & \textbf{90.4} & 88.5 \\
        \midrule
        \multicolumn{8}{l}{\textsc{\textbf{Question Answering}}} \\
        BoolQ & Acc. & 71.4 & 77.9 & \textbf{80.7} & 79.0 & 79.1 & {80.3} \\
        MultiRC & F1$_{\rm a}$ & 65.2 & 68.4 & 68.1 & \textbf{68.9} & 66.7 & 68.4 \\
        \midrule
        \textit{Average Score} & - & 80.4 & 82.8 & 83.5 & 83.5 & 83.7 & \textbf{84.0}\\
        \textit{\# Params Activated} & - & 110M & 110M & >166M & 110M & 110M & 110M \\
        \textit{\# Overall Params} & - & 110M & 110M & 450M & 231M & 240M & 231M \\
    \bottomrule
    \end{tabular}
    \caption{Results of all methods on 13 datasets. The best results are in \textbf{bold}. HMNet is our proposed method, and the subscript stands for three task relevance metrics, \ie, data property (d), manual design (md), and model-based relevance (m). }
    \vspace{-10px}
    \label{tab:overall}
\end{table*}

\section{Experiment}
\label{sec:exp_section}

\subsection{Datasets and Evaluation Metrics}
\label{sec:data_metric}
We conduct experiments on 13 NLU tasks and compare the performance of our HMNet with other baselines. These tasks can be grouped into four categories: 

(1) \textbf{Natural Language Inference (NLI):} These tasks aim to determine whether a \textit{hypothesis} is entailed, contradicted, or undetermined given a \textit{premise}. They require the model to measure the logical coherence between two sentences. We select six datasets: MNLI~\cite{williams2018broad}, QNLI, RTE, WNLI~\cite{wang-etal-2018-glue}, CB~\cite{DBLP:conf/nips/WangPNSMHLB19}, and SNLI~\cite{bowman-etal-2015-large}.

(2) \textbf{Sentiment Classification:} These tasks ask the model to classify the sentiment polarity of a sentence (\ie, positive or negative). We use IMDB~\cite{maas-etal-2011-learning} and SST-2~\cite{socher-etal-2013-recursive} datasets.

(3) \textbf{Similarity and Paraphrase:} These tasks aim at determining whether two sentences have the same or similar meaning. The semantic relationship between two sentences is vital in these tasks. We choose three commonly used datasets: QQP~\cite{wang-etal-2018-glue}, STS-B~\cite{cer-etal-2017-semeval}, and MRPC~\cite{dolan-brockett-2005-automatically}.

(4) \textbf{Question Answering:} This task requires the model to answer a question by reasoning on a given paragraph. BoolQ~\cite{clark-etal-2019-boolq} and MultiRC~\cite{khashabi-etal-2018-looking} datasets are used.

Detailed statistics of benchmark datasets are mentioned in Appendix~\ref{app:dataset_stat}.

\subsection{Baseline}
We compare with our HMNet with three other training methods:

\textbf{Single-Task fine-tuning:} We fine-tune a pre-trained BERT for each task independently. As a result, 13 models are obtained in total.

\textbf{Multi-Task fine-tuning:} Following MT-DNN~\cite{liu2019multi}, we use the pre-trained BERT as a shared encoder and add a task-special head for each task. During fine-tuning, all parameters are optimized by all tasks jointly. With multi-task learning, only one model needs to be trained and saved.

\textbf{SkillNet-style fine-tuning:} This is a sparsely activated multi-task learning structure proposed by~\citet{zhang2022skillnetnlu}. They design seven basic skills, such as getting the semantic meaning of a sequence and understanding a question. Then, they divide typical NLU tasks into five categories, each requiring a unique skill combination. The skill module is implemented by adding multiple FFN layers into the original Transformer structure. This model is also initialized by a pre-trained BERT. However, as each task activates multiple skill modules, compared with our method and other baselines, on average 1.0$\times$ more parameters (\ie, 223M vs. 110M) are used in inference. More detail about our reproduce reported in Appendix~\ref{app:skillnet_setting}.

\subsection{Implementation details}
For a fair comparison, all methods are initialized by the \texttt{bert-base-uncased} checkpoint. The batch size is 32, the max length of sequence is 512, and the initial learning rate is 2e-5, which is linearly decayed. AdamW~\cite{DBLP:conf/iclr/LoshchilovH19} optimizer is applied. We train all methods for three epochs. 

For the NLI and text similarity \& paraphrase tasks, similar to the vanilla BERT, we concatenate the sentence pair by adding a separator token \texttt{[SEP]} and a head token \texttt{[CLS]}. For the sentiment classification task, the input is a single sentence, so we only add a head token \texttt{[CLS]}.
% we take the vector of the \textit{[CLS]} token as the semantic representation of the sentence and obtain the probability of different emotions through linear and softmax. 
For the QA task, we concatenate the question, passage, and answer, then separate them by two \texttt{[SEP]} tokens. A head token \texttt{[CLS]} is also added at the beginning of the sequence.
% and judged whether the answer was the correct answer through the probability obtained by vector of the \textit{[CLS]} token after linear and softmax. 
All tasks are performed in similar ways, namely processing the embedding of the \texttt{[CLS]} token by a linear layer with a softmax activation function, and output the probability of each category.
The default setting of HMNet has eight shared layers, two task-clustering layers, and two task-specific layers.

\subsection{Experimental Results}\label{sec:result}
Table~\ref{tab:overall} shows the results on all datasets. 
In general, our HMNet performs better than other baselines on most datasets and achieves the best result in terms of the average score. This clearly demonstrates the superiority of our proposed HMNet. We further have the following observations:

(1) Comparing the performance between single-task and multi-task learning, it is evident that the latter can bring improvement for most tasks. Furthermore, this enhancement is related to the size of the datasets. Specifically, for the large dataset, such as MNLI and QQP (both of which have more than 300k training samples), the single-task learning can perform slightly better than multi-task learning. This implies that not all tasks can complement each other, and for those tasks with sufficient training data, adding extra tasks may not improve or even degrade performance. The potential reason is that the data in other tasks are collected from different domains and require models with distinct capabilities. Simply combining them will result in noise. For QA tasks, multi-task learning often leads to better performance. Indeed, QA is more complicated than other tasks, thus training on other tasks can be beneficial (\eg, better capture the relationship between question and answer). 

(2) Sparsely activated (SkillNet-style) methods can achieve better performance than single-/multi-task learning. Its advantages stem from two perspectives: First, this method groups tasks according to their requiring skills, so the tasks rely on similar skills can enhance one another (\eg, both NLI and semantic similarity judgment rely on the skill of understanding how two text segments interact), while avoiding noise from other used skills. On the other hand, since multiple skill modules are activated, more parameters improve the model's capacity. 

(3) Our HMNet with three different task relevance measurements can consistently outperform all baseline methods. Specifically, HMNet brings more than 0.7\% absolute improvement over multi-task learning in terms of average score. We attribute this improvement to our architecture of task-clustering and task-specific layers. Instead of sharing all layers, HMNet gradually separates the tasks in higher layers, so that the general language knowledge can be accumulated while the noise can be filtered out. Different from SkillNet-style methods, our HMNet only activates one channel at each layer, so its number of parameters is identical to a vanilla BERT. Surprisingly, HMNet with fewer parameters can even perform better. This demonstrates again the effectiveness of our proposed task clustering and the hierarchical sharing structure.

\subsection{Further Analysis}
\begin{figure}
    \centering
    \includegraphics[width=.9\linewidth]{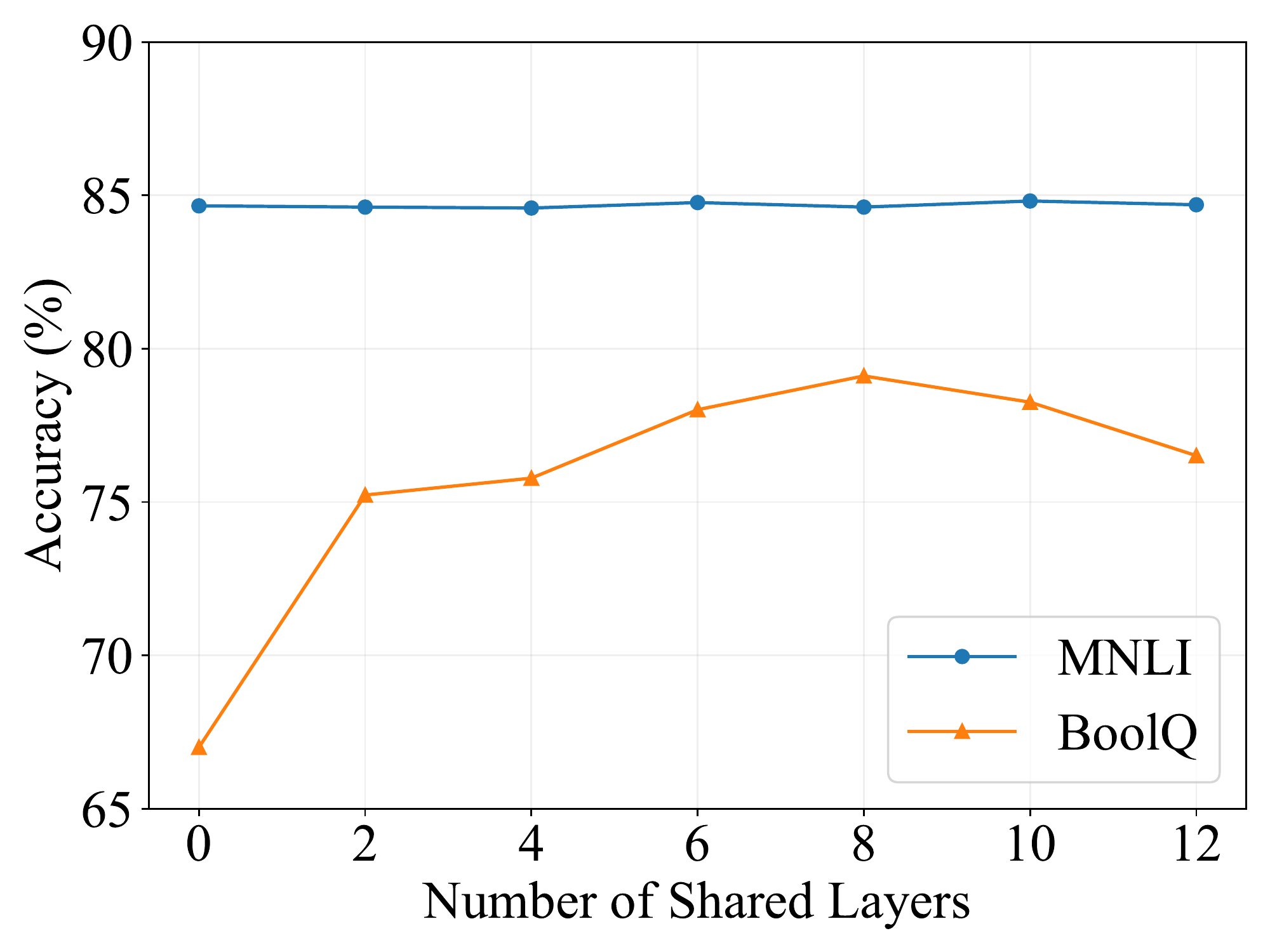}
    \caption{Influence of different numbers of shared layers in HMNet.}
    \vspace{-10px}
    \label{fig:shared}
\end{figure}
\paragraph{Influence of Shared Layers}
In HMNet, we design the bottom layers as fully shared, whilst the higher layers are only shared by task clusters or are task-specific. This approach is motivated by the hypothesis described in Section~\ref{sec:task_relevance}, which argues that the positive influence between tasks comes from capturing more general language knowledge at the bottom layer, and the negative impact comes from the noise brought by task interaction in the middle and top layers.  
% recent studies~\cite{jawahar2019does} on pre-trained language models, where the bottom layers are shown to capture general language knowledge. 
To prove this hypothesis, we experiment on using different numbers of shared layers to investigate their effect. This experiment is conducted on the MNLI and BoolQ datasets. We train HMNet on them and report their performance accordingly. Since only two tasks are considered, we categorize them into two clusters, so the task-clustering layers are transformed into task-specific. The results are shown in Figure~\ref{fig:shared}. 
We can observe that the performance of MNLI has minor changes as it contains sufficient training data. On the contrary, the result of BoolQ increases significantly when shared layers are used (from zero to six). This reflects the advantage of multi-task learning. However, when more shared layers are employed (more than six), the performance degrades. This confirms our assumption that task-specific knowledge is often learned in the upper layers, which supports our design of gradually separating tasks from the bottom up.

% Another conclusion of this work is that the multi-task fine-tuning can be effectively improved by sharing different task groups with different correlations at different levels of the model.  As shown in table x, the diversity of the structure also affects the performance of hierarchical multi-task training, where the annotation of $\{x, y, z\}$ represent different numbers of transformer layer at different levels of the model. Comparing different structures, we found that the model works best under the settings of 4-layer Share Layer, 6-layer Task-clustering Layer, and 2-layer Task-specific Layer. One of the explanations~\cite{rogers-etal-2020-primer} is that the lower layers of the model tend to learn the semantics of word-level and learn more syntactic information at the mid-layers, the information the final layer processed is usually task-specific. We compare the similarity of attention maps between the model fine-tuned by single-task and multi-task in~\Cref{app:atten_sim}. We find that the difference between the two models is less in the bottom layer and more in the upper layer. This also proves the rationality of hierarchical sharing to a certain extent.

\begin{figure}[t]
    \centering
    \includegraphics[width=.9\linewidth]{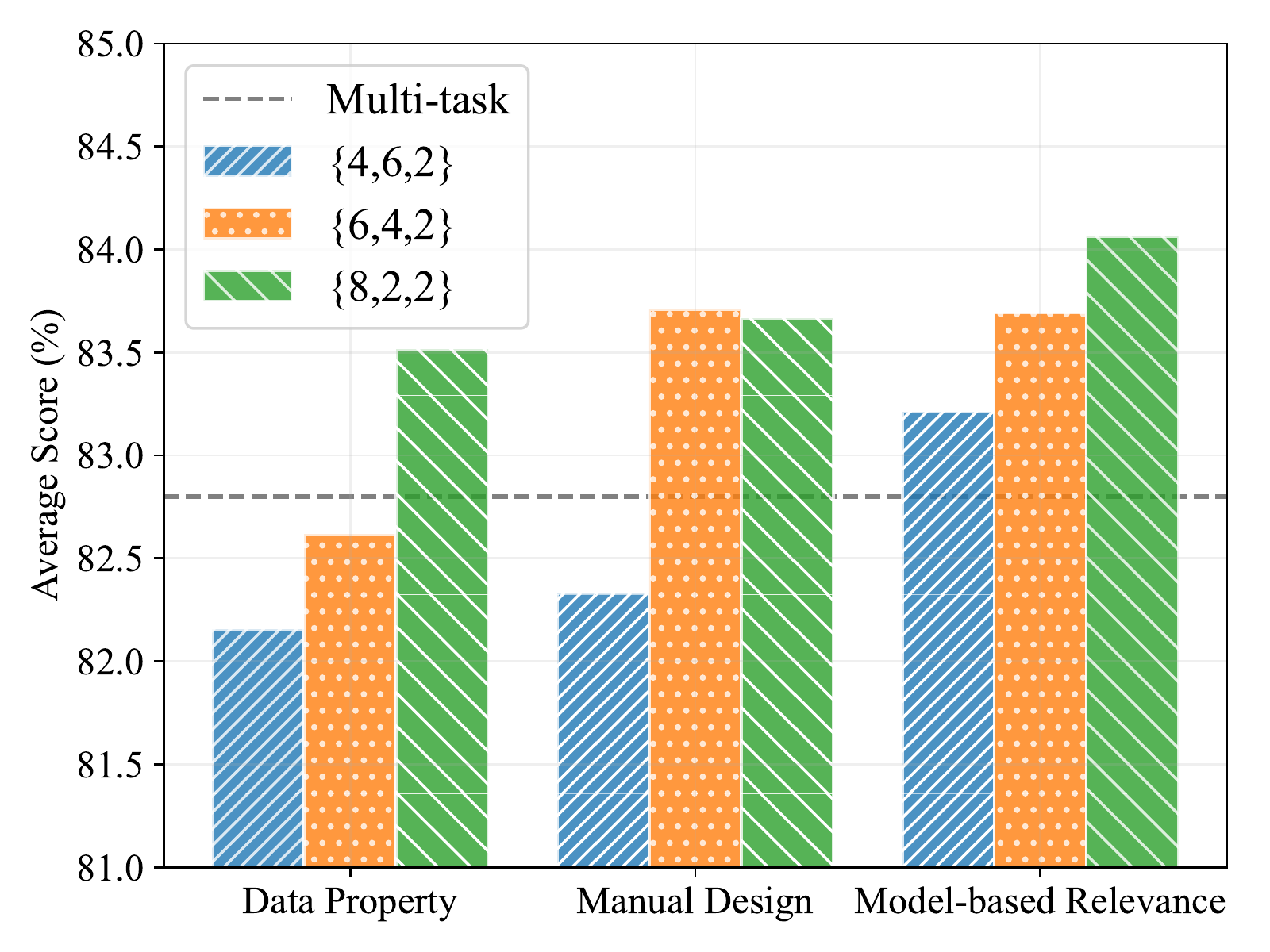}
    \caption{Results of average score with respect to different layer combinations.}
    \vspace{-10px}
    \label{fig:comb}
\end{figure}

\paragraph{Different Metrics for Task Relevance}
In Section~\ref{sec:task_relevance}, we devise three different metrics for task relevance, \ie, data property, manual design, and model-based relevance. Their performance is shown in Figure~\ref{fig:comb} and the right side of Table~\ref{tab:overall}. We can see: 
First, HMNet can outperform other baselines with any task relevance assessment. This highlights the significance of task clustering in multi-task learning. Moreover, the result demonstrates the adaptability of our method with regard to the evaluation of task relevance. As an early exploration of the effect of task relevance on multi-task learning, the three metrics we have provided are very preliminary. We believe that a more accurate task relevance could bring further improvement. Second, the model-based similarity performs the best. With the help of deep neural models, the task relevance in high dimensions can be better captured. Such relevance is hard to be observed by humans or extracted from shallow data properties. In addition, quantifying the tasks' relevance from the perspective of the model can narrow the gap between task clustering and multi-task learning. Notably, though better performance is obtained, the model-based metric needs additional cost on model training. The other two metrics can avoid it.

\paragraph{Influence of Different Layer Combinations}
HMNet has three different kinds of layers, so their different combinations may influence the performance. We conduct an experiment by tuning the number of layers at each level. For clarity, we use $\{x, y, z\}$ to denote a HMNet with $x$ shared layers, $y$ task-clustering layers, and $z$ task-specific layers. Experimental results are shown in Figure~\ref{fig:comb}. Using eight shared layers yields the optimal performance. In comparison to traditional multi-task learning that shares all layers, HMNet with more than six shared layers can obtain better results. This implies that sharing all layers is not an effective strategy for multi-task learning, and our hierarchical sharing structure can mitigate the negative effect across tasks. Besides, when equipping with the model-based similarity, HMNet outperforms multi-task learning with any combination of shared, task-clustering, and task-specific layers. This reflects the robustness of our method and validates that the excellent performance of our method stems from our HMNet architecture and the consideration of task relevance rather than finely tuned hyperparameters.

\section{Conclusion and Future Work}
In this work, we explored task correlation and built a hierarchical multi-task learning framework. Our framework adopts a coarse-to-fine manner, in which the tasks are gradually separated from the bottom up. By doing so, it can reap the benefits of multi-tasking learning at the lower layers while avoiding its harmful impact on the upper layers. Extensive experiments on several challenging NLU datasets showed that our model achieves better performance than existing strategies. Further experiments indicated that our methods are flexible with the choice of task relevance metrics, and robust with the hyperparameter selection. As a preliminary study on incorporating task relevance into multi-task learning, there are several potential future directions, such as new backbone models and task relevance metrics.

\bibliography{custom}

\appendix
% \section{Example Appendix}
% \label{sec:appendix}

% This is an appendix.
% \section{Dataset Statistics}
% \label{app:dataset_stat}
% We statistic the datasets used in this paper in Table~\ref{tab:datasets}.

%\begin{table}[htbp]
%  \centering
%  \small
%    \begin{tabular}{llrr}
%    \toprule
%    Category & Datasets & Train & Dev \\
%    \midrule
%    \multirow{6}{*}{NLI} & MNLI & 392k & 19k \\
%    & QNLI & 104k & 5.5k\\
%    & RTE & 2.5k & 277 \\
%    & WNLI & 635 & 71\\
%    & CB & 250 & 56 \\
%    & SNLI & 54k & 9.8k \\
%     \midrule
%     \multirow{2}{*}{Sentiment Classification} & IMDB & 25k &  25k \\
%     & SST-2 & 67k & 872 \\
%     \midrule
%     \multirow{3}{*}{Text Similariy \& Paraphrase} & QQP & 363k & 4k \\
%     & STS-B & 5.7k & 1.5k \\
%     & MRPC & 3.6k & 408 \\
%     \midrule
%     \multirow{2}{*}{Question Answering}& BooLQ & 9.4k & 3.2k \\
%     & MultiRC & 27k & 4.8k \\
%     \bottomrule
%     \end{tabular}
%     \caption{The statistics of datasets.}
%     \label{tab:datasets_stat}
% \end{table}%

% \section{The Clustering Results of Data Property}
% \label{app:text_relevance}

% According to Section~\ref{sec:text_corrl}

\clearpage

\section{The Clustering Results}
\label{app:task_relevance}

\begin{table*}[htbp]
  \centering
  \tiny
    \begin{tabular}{cccccccccccccc}
    \toprule
          & \multicolumn{1}{l}{\textbf{MNLI}} & \multicolumn{1}{l}{\textbf{QNLI}} & \multicolumn{1}{l}{\textbf{RTE}} & \multicolumn{1}{l}{\textbf{WNLI}} & \multicolumn{1}{l}{\textbf{CB}} & \multicolumn{1}{l}{\textbf{SNLI}} & \multicolumn{1}{l}{\textbf{IMDB}} & \multicolumn{1}{l}{\textbf{SST-2}} & \multicolumn{1}{l}{\textbf{QQP}} & \multicolumn{1}{l}{\textbf{STSB}} & \multicolumn{1}{l}{\textbf{MRPC}} & \multicolumn{1}{l}{\textbf{BoolQ}} & \multicolumn{1}{l}{\textbf{MultiRC}} \\
    \midrule
    \textbf{MNLI} & 100.00\% & 92.83\% & 97.64\% & 99.58\% & 99.42\% & 98.50\% & 95.70\% & 99.21\% & 93.46\% & 98.29\% & 98.21\% & 93.83\% & 98.76\% \\
    \textbf{QNLI} & 94.40\% & 100.00\% & 98.11\% & 98.32\% & 98.01\% & 95.02\% & 94.00\% & 96.55\% & 94.19\% & 97.71\% & 98.02\% & 95.31\% & 97.55\% \\
    \textbf{RTE} & 52.42\% & 51.80\% & 100.00\% & 84.72\% & 81.73\% & 59.25\% & 52.99\% & 65.23\% & 52.26\% & 74.24\% & 74.11\% & 54.11\% & 66.54\% \\
    \textbf{WNLI} & 5.50\% & 5.34\% & 8.72\% & 100.00\% & 24.80\% & 7.47\% & 5.63\% & 9.90\% & 5.51\% & 10.64\% & 9.77\% & 5.75\% & 8.78\% \\
    \textbf{CB} & 10.62\% & 10.29\% & 16.26\% & 47.93\% & 100.00\% & 13.76\% & 10.87\% & 18.86\% & 10.61\% & 18.95\% & 18.44\% & 11.00\% & 16.15\% \\
    \textbf{SNLI} & 69.92\% & 66.33\% & 78.35\% & 95.94\% & 91.44\% & 100.00\% & 70.87\% & 84.65\% & 68.98\% & 83.62\% & 80.12\% & 69.19\% & 80.05\% \\
    \textbf{IMDB} & 93.74\% & 90.55\% & 96.68\% & 99.72\% & 99.67\% & 97.80\% & 100.00\% & 99.69\% & 91.86\% & 97.45\% & 97.06\% & 92.13\% & 97.41\% \\
    \textbf{SST-2} & 44.45\% & 42.54\% & 54.44\% & 80.31\% & 79.12\% & 53.43\% & 45.60\% & 100.00\% & 44.07\% & 58.34\% & 57.02\% & 44.68\% & 56.31\% \\
    \textbf{QQP} & 92.93\% & 92.11\% & 96.79\% & 99.09\% & 98.80\% & 96.63\% & 93.25\% & 97.80\% & 100.00\% & 97.67\% & 97.40\% & 93.65\% & 97.13\% \\
    \textbf{STSB} & 41.08\% & 40.17\% & 57.80\% & 80.52\% & 74.18\% & 49.24\% & 41.59\% & 54.43\% & 41.06\% & 100.00\% & 67.28\% & 42.43\% & 54.52\% \\
    \textbf{MRPC} & 43.97\% & 43.15\% & 61.80\% & 79.12\% & 77.30\% & 50.53\% & 44.36\% & 56.97\% & 43.85\% & 72.06\% & 100.00\% & 45.47\% & 57.67\% \\
    \textbf{BoolQ} & 87.67\% & 87.58\% & 94.18\% & 97.20\% & 96.27\% & 91.07\% & 87.88\% & 93.19\% & 88.01\% & 94.84\% & 94.90\% & 100.00\% & 93.29\% \\
    \textbf{MultiRC} & 57.48\% & 55.84\% & 72.14\% & 92.43\% & 88.03\% & 65.64\% & 57.88\% & 73.15\% & 56.86\% & 75.92\% & 74.98\% & 58.11\% & 100.00\% \\
    \bottomrule
    \end{tabular}
    \caption{\textbf{The task relevance according data property.} This task relevance will be measured by the vocabulary co-occurrence between the task pair, the formulation as Equation(\ref{eq:text_corr}).}
    \label{tab:text_corr}
\end{table*}

\begin{table*}[ht]
  \centering
  \tiny
    \begin{tabular}{crrrrrrrrrrrrr}
    \toprule
          & \multicolumn{1}{c}{\textbf{MNLI}} & \multicolumn{1}{c}{\textbf{QNLI}} & \multicolumn{1}{c}{\textbf{RTE}} & \multicolumn{1}{c}{\textbf{WNLI}} & \multicolumn{1}{c}{\textbf{CB}} & \multicolumn{1}{c}{\textbf{SNLI}} & \multicolumn{1}{c}{\textbf{IMDB}} & \multicolumn{1}{c}{\textbf{SST-2}} & \multicolumn{1}{c}{\textbf{QQP}} & \multicolumn{1}{c}{\textbf{STS-B}} & \multicolumn{1}{c}{\textbf{MRPC}} & \multicolumn{1}{c}{\textbf{BoolQ}} & \multicolumn{1}{c}{\textbf{MultiRC}} \\
    \midrule
    \textbf{MNLI} & 0.00\% & \cellcolor[rgb]{ .949,  .961,  .98}-0.10\% & \cellcolor[rgb]{ .855,  .588,  .58}15.96\% & \cellcolor[rgb]{ .902,  .714,  .71}50.00\% & \cellcolor[rgb]{ .855,  .588,  .58}47.31\% & \cellcolor[rgb]{ .855,  .588,  .58}0.75\% & \cellcolor[rgb]{ .969,  .976,  .988}-0.29\% & 0.00\% & \cellcolor[rgb]{ .639,  .741,  .863}-0.14\% & \cellcolor[rgb]{ .753,  .824,  .906}-0.96\% & \cellcolor[rgb]{ .753,  .82,  .906}-1.87\% & \cellcolor[rgb]{ .855,  .588,  .58}5.93\% & \cellcolor[rgb]{ .855,  .588,  .58}6.42\% \\
    \textbf{QNLI} & \cellcolor[rgb]{ .706,  .788,  .886}-0.54\% & 0.00\% & \cellcolor[rgb]{ .984,  .949,  .945}2.13\% & \cellcolor[rgb]{ .957,  .878,  .875}21.43\% & \cellcolor[rgb]{ .922,  .773,  .769}26.37\% & \cellcolor[rgb]{ .957,  .875,  .871}0.24\% & \cellcolor[rgb]{ .988,  .992,  .996}-0.09\% & \cellcolor[rgb]{ .882,  .914,  .953}-0.62\% & \cellcolor[rgb]{ .835,  .882,  .937}-0.06\% & \cellcolor[rgb]{ .949,  .851,  .851}0.43\% & \cellcolor[rgb]{ .906,  .933,  .965}-0.69\% & \cellcolor[rgb]{ .871,  .627,  .62}5.38\% & \cellcolor[rgb]{ .965,  .902,  .898}1.58\% \\
    \textbf{RTE} & \cellcolor[rgb]{ .745,  .816,  .902}-0.47\% & \cellcolor[rgb]{ .835,  .882,  .937}-0.32\% & 0.00\% & \cellcolor[rgb]{ .988,  .961,  .961}7.14\% & \cellcolor[rgb]{ .969,  .91,  .91}10.45\% & \cellcolor[rgb]{ .953,  .867,  .863}0.25\% & \cellcolor[rgb]{ .855,  .588,  .58}0.10\% & \cellcolor[rgb]{ .812,  .867,  .929}-0.98\% & \cellcolor[rgb]{ .929,  .949,  .973}-0.03\% & \cellcolor[rgb]{ .973,  .98,  .988}-0.10\% & \cellcolor[rgb]{ .741,  .816,  .902}-1.95\% & \cellcolor[rgb]{ .984,  .957,  .953}0.67\% & \cellcolor[rgb]{ .98,  .937,  .933}1.03\% \\
    \textbf{WNLI} & \cellcolor[rgb]{ .847,  .89,  .941}-0.28\% & \cellcolor[rgb]{ .835,  .882,  .937}-0.32\% & \cellcolor[rgb]{ .584,  .702,  .843}-3.19\% & 0.00\% & \cellcolor[rgb]{ .992,  .98,  .98}2.62\% & \cellcolor[rgb]{ .988,  .961,  .957}0.08\% & \cellcolor[rgb]{ .89,  .678,  .675}0.08\% & \cellcolor[rgb]{ .859,  .898,  .945}-0.74\% & \cellcolor[rgb]{ .922,  .941,  .969}-0.03\% & \cellcolor[rgb]{ .957,  .969,  .98}-0.16\% & \cellcolor[rgb]{ .945,  .961,  .98}-0.39\% & \cellcolor[rgb]{ .988,  .961,  .961}0.59\% & \cellcolor[rgb]{ .714,  .792,  .89}-0.18\% \\
    \textbf{CB} & \cellcolor[rgb]{ .808,  .863,  .925}-0.35\% & \cellcolor[rgb]{ .929,  .796,  .792}0.02\% & 0.00\% & \cellcolor[rgb]{ .988,  .961,  .961}7.14\% & 0.00\% & \cellcolor[rgb]{ .584,  .702,  .843}-0.30\% & \cellcolor[rgb]{ .988,  .992,  .996}-0.09\% & \cellcolor[rgb]{ .906,  .933,  .965}-0.49\% & \cellcolor[rgb]{ .886,  .678,  .671}0.04\% & \cellcolor[rgb]{ .961,  .878,  .878}0.35\% & \cellcolor[rgb]{ .933,  .949,  .973}-0.50\% & \cellcolor[rgb]{ .584,  .702,  .843}-0.17\% & \cellcolor[rgb]{ .996,  .984,  .984}0.29\% \\
    \textbf{SNLI} & \cellcolor[rgb]{ .894,  .925,  .961}-0.19\% & \cellcolor[rgb]{ .675,  .765,  .875}-0.64\% & \cellcolor[rgb]{ .894,  .698,  .694}11.70\% & \cellcolor[rgb]{ .871,  .631,  .624}64.29\% & \cellcolor[rgb]{ .898,  .71,  .706}33.51\% & 0.00\% & \cellcolor[rgb]{ .98,  .988,  .992}-0.17\% & \cellcolor[rgb]{ .651,  .749,  .867}-1.85\% & \cellcolor[rgb]{ .686,  .776,  .882}-0.12\% & \cellcolor[rgb]{ .686,  .773,  .878}-1.23\% & \cellcolor[rgb]{ .584,  .702,  .843}-3.16\% & \cellcolor[rgb]{ .925,  .78,  .776}3.19\% & \cellcolor[rgb]{ .929,  .8,  .796}3.16\% \\
    \textbf{IMDB} & \cellcolor[rgb]{ .992,  .976,  .976}0.01\% & 0.00\% & \cellcolor[rgb]{ .859,  .898,  .945}-1.06\% & \cellcolor[rgb]{ .973,  .918,  .918}14.29\% & \cellcolor[rgb]{ .906,  .733,  .725}30.95\% & \cellcolor[rgb]{ .953,  .965,  .98}-0.03\% & 0.00\% & \cellcolor[rgb]{ .859,  .898,  .945}-0.74\% & \cellcolor[rgb]{ .584,  .702,  .843}-0.16\% & \cellcolor[rgb]{ .937,  .824,  .82}0.52\% & \cellcolor[rgb]{ .902,  .718,  .714}0.79\% & \cellcolor[rgb]{ .992,  .976,  .976}0.38\% & \cellcolor[rgb]{ .988,  .957,  .957}0.68\% \\
    \textbf{SST-2} & \cellcolor[rgb]{ .855,  .588,  .58}0.21\% & \cellcolor[rgb]{ .988,  .992,  .996}-0.02\% & \cellcolor[rgb]{ .992,  .976,  .973}1.06\% & \cellcolor[rgb]{ .973,  .918,  .918}14.29\% & \cellcolor[rgb]{ .898,  .71,  .706}33.58\% & \cellcolor[rgb]{ .984,  .953,  .953}0.09\% & \cellcolor[rgb]{ .584,  .702,  .843}-4.24\% & 0.00\% & \cellcolor[rgb]{ .941,  .827,  .824}0.02\% & \cellcolor[rgb]{ .804,  .859,  .925}-0.76\% & \cellcolor[rgb]{ .737,  .812,  .898}-1.98\% & \cellcolor[rgb]{ .886,  .675,  .671}4.71\% & \cellcolor[rgb]{ .988,  .969,  .969}0.53\% \\
    \textbf{QQP} & \cellcolor[rgb]{ .584,  .702,  .843}-0.77\% & \cellcolor[rgb]{ .584,  .702,  .843}-0.82\% & \cellcolor[rgb]{ .973,  .922,  .918}3.19\% & \cellcolor[rgb]{ .584,  .702,  .843}0.00\% & \cellcolor[rgb]{ .98,  .937,  .937}7.48\% & \cellcolor[rgb]{ .984,  .953,  .953}0.09\% & \cellcolor[rgb]{ .957,  .969,  .984}-0.41\% & \cellcolor[rgb]{ .584,  .702,  .843}-2.21\% & 0.00\% & \cellcolor[rgb]{ .584,  .702,  .843}-1.63\% & \cellcolor[rgb]{ .624,  .729,  .855}-2.85\% & \cellcolor[rgb]{ .957,  .871,  .867}1.89\% & \cellcolor[rgb]{ .686,  .773,  .878}-0.19\% \\
    \textbf{STS-B} & \cellcolor[rgb]{ .933,  .953,  .973}-0.12\% & \cellcolor[rgb]{ .855,  .588,  .58}0.04\% & \cellcolor[rgb]{ .949,  .851,  .847}5.85\% & \cellcolor[rgb]{ .886,  .675,  .667}57.14\% & \cellcolor[rgb]{ .961,  .886,  .886}13.07\% & \cellcolor[rgb]{ .976,  .929,  .925}0.13\% & \cellcolor[rgb]{ .957,  .878,  .875}0.03\% & \cellcolor[rgb]{ .769,  .831,  .91}-1.23\% & \cellcolor[rgb]{ .855,  .588,  .58}0.05\% & 0.00\% & \cellcolor[rgb]{ .855,  .588,  .58}1.15\% & \cellcolor[rgb]{ .925,  .784,  .78}3.15\% & \cellcolor[rgb]{ .969,  .906,  .906}1.48\% \\
    \textbf{MRPC} & \cellcolor[rgb]{ .886,  .918,  .957}-0.21\% & \cellcolor[rgb]{ .929,  .796,  .792}0.02\% & \cellcolor[rgb]{ .976,  .933,  .933}2.66\% & \cellcolor[rgb]{ .957,  .878,  .875}21.43\% & \cellcolor[rgb]{ .961,  .886,  .886}13.07\% & \cellcolor[rgb]{ .984,  .988,  .992}-0.01\% & \cellcolor[rgb]{ .996,  .996,  .996}-0.01\% & \cellcolor[rgb]{ .855,  .588,  .58}0.25\% & \cellcolor[rgb]{ .718,  .796,  .89}-0.11\% & \cellcolor[rgb]{ .941,  .957,  .976}-0.23\% & 0.00\% & \cellcolor[rgb]{ .957,  .878,  .878}1.77\% & \cellcolor[rgb]{ .996,  .988,  .988}0.22\% \\
    \textbf{BoolQ} & \cellcolor[rgb]{ .808,  .863,  .925}-0.35\% & \cellcolor[rgb]{ .816,  .867,  .929}-0.36\% & \cellcolor[rgb]{ .651,  .749,  .867}-2.66\% & \cellcolor[rgb]{ .957,  .878,  .875}21.43\% & \cellcolor[rgb]{ .898,  .71,  .706}33.58\% & \cellcolor[rgb]{ .796,  .855,  .922}-0.15\% & \cellcolor[rgb]{ .992,  .996,  .996}-0.05\% & \cellcolor[rgb]{ .882,  .914,  .953}-0.62\% & \cellcolor[rgb]{ .914,  .937,  .965}-0.03\% & \cellcolor[rgb]{ .855,  .588,  .58}1.18\% & \cellcolor[rgb]{ .835,  .882,  .937}-1.24\% & 0.00\% & \cellcolor[rgb]{ .584,  .702,  .843}-0.26\% \\
    \textbf{MultiRC} & \cellcolor[rgb]{ .796,  .855,  .922}-0.37\% & \cellcolor[rgb]{ .898,  .925,  .961}-0.20\% & \cellcolor[rgb]{ .965,  .894,  .89}4.26\% & \cellcolor[rgb]{ .855,  .588,  .58}71.43\% & \cellcolor[rgb]{ .969,  .91,  .91}10.38\% & \cellcolor[rgb]{ .906,  .933,  .965}-0.07\% & \cellcolor[rgb]{ .996,  .996,  .996}-0.04\% & \cellcolor[rgb]{ .722,  .8,  .894}-1.48\% & \cellcolor[rgb]{ .902,  .722,  .714}0.04\% & \cellcolor[rgb]{ .918,  .757,  .753}0.70\% & \cellcolor[rgb]{ .969,  .906,  .902}0.27\% & \cellcolor[rgb]{ .961,  .886,  .886}1.64\% & 0.00\% \\
    \bottomrule
    \end{tabular}%
    \caption{\textbf{The model-based task relevance} The task relevance is defined by the increased ratio by the performance of single-task tuning relative to the performance of co-training between source task with target task, and the formulation as Equation (\ref{eq:task_corr}).}
    \label{tab:task_corr}%
\end{table*}%

\begin{table*}[t!]
    \centering
    \small
    \begin{tabular}{lccccc}
    \toprule
         \textbf{Corpus} & \#Train & \#Dev & \#Test & \#Label & Metrics  \\
    \midrule
      \multicolumn{5}{l}{\textsc{\textbf{Natural Language Inference}}} \\
       MNLI & 393K & 20k & 20k & 3 & Accuracy \\
       QNLI & 108k & 5.7k & 5.7k & 2 & Accuracy \\
       RTE & 2.5k & 276 & 3k & 2 & Accuracy \\
       WNLI & 634 & 71 & 146 & 2 & Accuracy \\
       CB & 250 & 57 & 250 & 2 & Accuracy/F1 \\
       SNLI & 549k & 9.8k & 9.8k & 3 & Accuracy \\
        \midrule
        \multicolumn{5}{l}{\textbf{\textsc{Sentiment Classification}}} \\
        IMDB & 25k & 0k & 25k & 2 & Accuracy \\
        SST-2 & 67K & 872 & 1.8k & 2 & Accuracy \\
       \midrule
        \multicolumn{5}{l}{\textsc{\textbf{Similarity and Paraphrase}}} \\
        QQP & 364k & 40k & 391k & 2 & Accuracy/F1 \\
        STS-B & 7K & 1.5k & 1.4k & 1 & Pearson/Spearman corr \\
        MRPC & 3.7k & 408 & 1.7k & 2 & Accuracy/F1 \\
        \midrule
        \multicolumn{5}{l}{\textsc{\textbf{Question Answering}}} \\
        BoolQ & 9.4k & 3.3k & 3.2k & 2 & Accuracy \\
        MultiRC & 5.1k & 953 & 1.8k & 2 & F1$_{\rm a}$/EM \\
    \bottomrule
    \end{tabular}
    \caption{Summary of the 13 datasets.}
    \label{tab:datasets}
\end{table*}

% \subsection{Text correlation}\label{sec:text_corr}

% We group tasks based on text correlation between them. The proportion of the intersection of two datasets' vocabulary to the original vocabulary is defined as the textual correlation between two datasets, and the formulation is as follows:

% \begin{equation}
%     s_{text\ correlation} = (\mathbb{V}_{source} \cap \mathbb{V}_{target}) / \mathbb{V}_{target}.
%     \label{eq:text_corr}
% \end{equation}

% \Cref{tab:text_corr} shows the textual connection between different datasets. Then, as a feature, we divide these tasks into three categories based on their task correlation. The grouping result is shown in~\Cref{fig:text_corr}.

As described in Section~\ref{sec:task_relevance}, we group these tasks from three prospects, individually data property, manual design, and model-based relevance. As for the data property of datasets, Table~\ref{tab:text_corr} is the full table which reports the task relevance of all 13 datasets. And using these features, we cluster these tasks into three groups using $k$-means. The result is \{WNLI, CB\}, \{MultiRC, RTE, SST-2, MRPC, STS-B\}, \{QQP, QNLI, BoolQ, IMDB, SNLI, MNLI\}.

As for task relevance of model based, we report this relevance of all tasks in Table~\ref{tab:task_corr}. As described before, we cluster these tasks using $k$-means according to this relevance as feature. The results is \{CB, WNLI, QQP, RTE\}, \{MRPC, QNLI, BoolQ, IMDB, SST-2\}, \{MultiRC, STS-B, SNLI, MNLI\}.

% \begin{figure}[!ht]
%     \centering
%     \includegraphics[width=0.45\textwidth]{fig/kmeans_dist_3.eps}
%     \caption{The grouping results according to task relevance based on data property.}
%     \label{fig:text_corr}
% \end{figure}

% \begin{figure}[!ht]
%     \centering
%     \includegraphics[width=0.45\textwidth]{fig/kmeans_corel_3.eps}
%     \caption{The grouping results according model based task relevance.}
%     \label{fig:task_corr}
% \end{figure}

\section{Dataset Statistics}
\label{app:dataset_stat}
We statistic the datasets used in this paper in Table~\ref{tab:datasets}.

\section{Skillnet Reproduce Setting}
\label{app:skillnet_setting}

\begin{table*}[!ht]
\small
\centering
\begin{tabular}{clllllllll}
\toprule
 & & \multicolumn{7}{c}{\textbf{Skills}} & \\
% \midrule
\textbf{ID} & \textbf{Task Category} & s1 & s2 & s3 & s4 & s5 & s6 & s7 & \textbf{Datasets} \\ \midrule
T1 & Sentiment Analysis & \checkmark & & & \checkmark & & & \checkmark & IMDB, SST-2\\
T2 & Natural Language Inference & \checkmark & & \checkmark & & & & \checkmark & MNLI, QNLI, \textit{etc}. \\
T3 & Semantic Similarity & \checkmark & & \checkmark & & & \checkmark & \checkmark & QQP, STS-B\\
T4 & Text Classification & \checkmark & & & & & & \checkmark & MRPC\\
T5 & Question Answering &  & \checkmark & \checkmark & & \checkmark & & \checkmark & BooLQ, MultiRC \\ \bottomrule
\end{tabular}
\caption{The different skill module will be activated for each task in SkillNet Fine-tuning, the activated skill is marked with a tick.}
\label{tab:skill}
\end{table*}

\begin{table}[!ht]
\small
\centering
\begin{tabular}{ll}
\toprule 
Skill  & Description  \\
\midrule 
s1  & get the semantic meaning of a sequence  \\
s2  & get the semantic meaning of a token  \\
s3  & understand how two text segments interact  \\
s4  & understand the sentiment of texts  \\
s5  & understand natural language questions  \\
s6  & understand texts in finance domain  \\
s7  & generic skill  \\
\bottomrule
\end{tabular}
\caption{The example of skills and description in Skillnet-style Fine-tuning~\cite{zhang2022skillnetnlu}}
\label{tab:skill_desc}
\end{table}

We reproduce SkillNet~\cite{zhang2022skillnetnlu} according to the paper since they does not release the code and models. We use the same skill setting and the method of skill module activated by different tasks, See Table~\ref{tab:skill_desc}, Table~\ref{tab:skill}.

\section{Attention Map Similarity}\label{app:atten_sim}

\begin{figure}[!ht]
    \centering
    % \vspace{-1cm}
    \includegraphics[width=\linewidth]{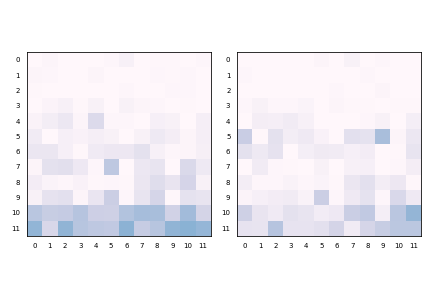}
    \caption{Per head attention maps' cosine similarity between fine-tuned model and co-trained model using positive task pair(QQP \& RTE). Darker colors means greater differences. left: QQP right: RTE}
    \label{fig:compare_same}
\end{figure}

\begin{figure}[!ht]
    \centering
    \includegraphics[width=0.5\textwidth]{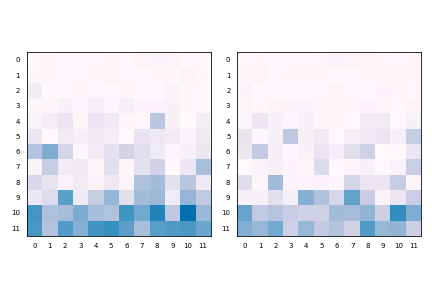}
    \caption{Per head attention maps' cosine similarity between fine-tuned model and co-trained model using negative task pair(MRPC \& RTE). Darker colors means greater differences. left: MRPC right: RTE}
    \label{fig:compare_diff}
\end{figure}

Following previous work~\cite{liu2019multi, rogers-etal-2020-primer}, we compare the attention maps between fine-tuned model and the co-trained model to explore the behavior during multi-task learning. As shown in Table~\ref{tab:task_corr}, the influence between RTE and QQP is relatively positive and between RTE and MRPC is relatively negative. We conduct some experiments using these tasks to explore the training difference between positive task-pair and negative pairs. Figure~\ref{fig:compare_same} compares the similarity between positive task pair(QQP \& RTE), and Figure~\ref{fig:compare_diff} compares the negative task-pair(MRPC \& RTE). As shown in these figures, both positive pair or negative, the attention map of the bottom six layers is very similar, and the top two layers are very different. Comparing Figure~\ref{fig:compare_same} and Figure~\ref{fig:compare_diff}, we can see the attention map in the positive pair is more similar in the middle layers. Therefore, when considering the model structure, we choose all tasks in the bottom layers that are completely shared to learn word-level information, the middle layers are used to learn semantic information, and the last two layers are used to learn task-level information.  

\end{document}